\documentclass[10pt,twocolumn,letterpaper]{article}

\usepackage{cvpr}              

\usepackage[dvipsnames]{xcolor}

\usepackage{xfrac}

\definecolor{cvprblue}{rgb}{0.21,0.49,0.74}
\usepackage[pagebackref,breaklinks,colorlinks,citecolor=cvprblue]{hyperref}
\usepackage{multirow}             
\usepackage{float}                  
\usepackage{subcaption}
\usepackage{overpic} 
\usepackage{caption}
\usepackage{wrapfig}

\def\paperID{6705} 
\def\confName{CVPR}
\def\confYear{2024}

\iftrue

\fi

\title{Tailored Visions: Enhancing Text-to-Image Generation with Personalized Prompt Rewriting}

\author{\vspace{-3mm}
\centerline{Zijie Chen$^{*,1,2}$ ~~ Lichao Zhang$^{*,2}$ ~~ Fangsheng Weng$^{3}$ ~~  Lili Pan$^{\dagger,4}$ ~~ Zhenzhong Lan$^{\dagger,2}$}\\\\ 
\centerline{\normalfont{$^1$Zhejiang University} \quad {$^2$Westlake University} \quad {$^{3}$Scietrain}}\\ 
\centerline{{$^4$University of Electronic Science and Technology of China}}\\
\small \centerline{\texttt{\{chenzijie, zhanglichao, lanzhenzhong\}@westlake.edu.cn , lilipan@uestc.edu.cn}}
}

\begin{document}
\maketitle
\renewcommand{\thefootnote}{$^*$}
\footnotetext{Equal contribution.}
\renewcommand{\thefootnote}{$^\dagger$}
\footnotetext{Corresponding author.}

\begin{abstract}
 Despite significant progress in the field, it is still challenging to create personalized visual representations that align closely with the desires and preferences of individual users. This process requires users to articulate their ideas in words that are both comprehensible to the models and accurately capture their vision, posing difficulties for many users. In this paper, we tackle this challenge by leveraging historical user interactions with the system to enhance user prompts. We propose a novel approach that involves rewriting user prompts based on a newly collected large-scale text-to-image dataset with over 300k prompts from 3115 users. Our rewriting model enhances the expressiveness and alignment of user prompts with their intended visual outputs. Experimental results demonstrate the superiority of our methods over baseline approaches, as evidenced in our new offline evaluation method and online tests. 
 {Our code and dataset are available at https://github.com/zzjchen/Tailored-Visions}
\end{abstract} 
\section{Introduction}
\label{sec:intro}

\begin{figure}[!th]
\begin{center}
\includegraphics[width=\columnwidth]{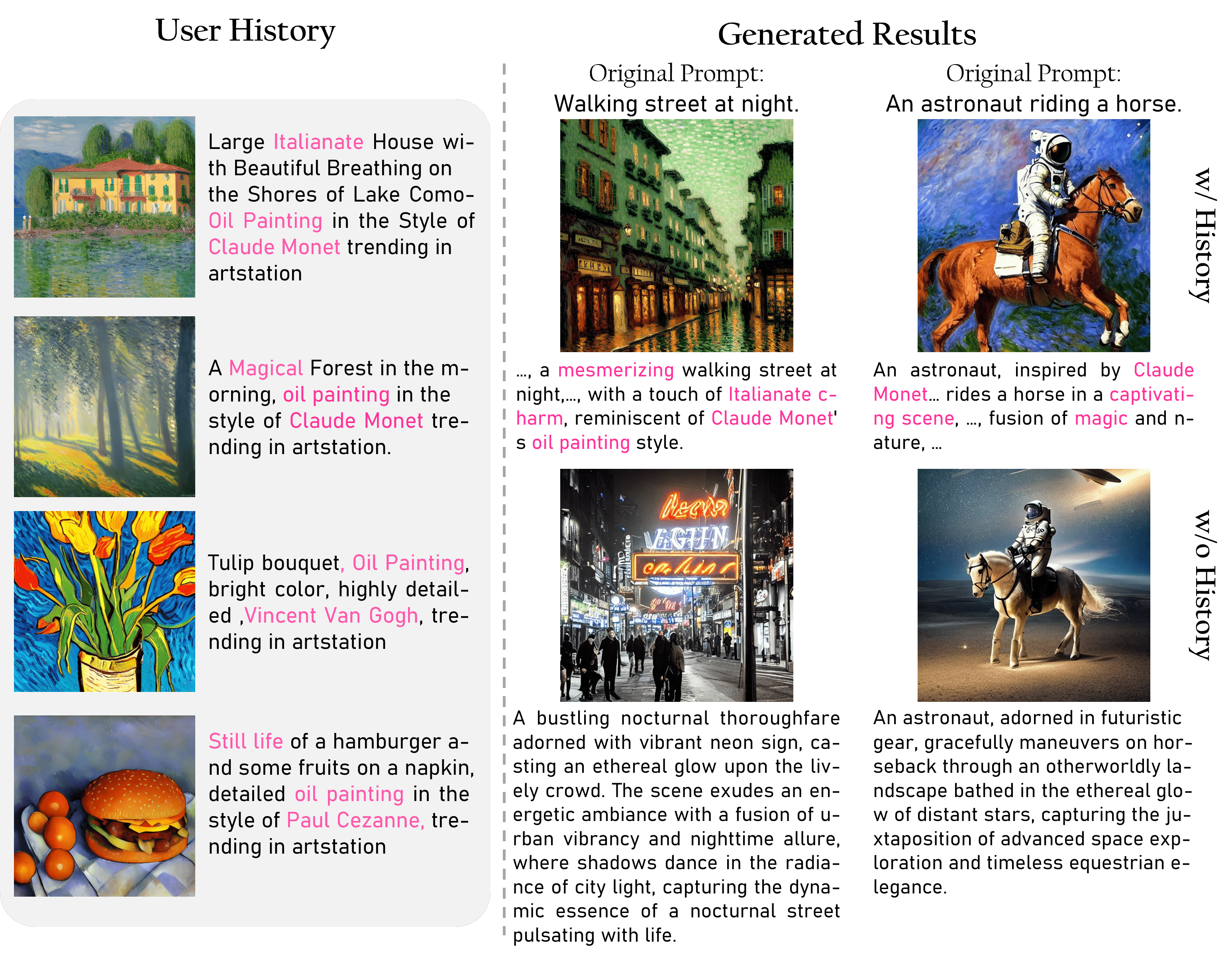}
\end{center}
\caption{Comparison between our personalized prompt rewriting method and the standard prompt rewriting method. Our technique excels at incorporating user preferences, such as "oil paintings by artists," while methods that lack a historical context frequently generate content that may not align with the user's desires. }
\label{fig: Fig1}
\end{figure}


Increasingly large and powerful foundation models~\citep{brown2020language,Rombach_2022_CVPR,gal2022image} are trained through self-supervised learning. These large pretrained models (LPMs) serve as efficient compressors \citep{deletang2023language}, condensing vast amounts of internet data. This compression enables the convenient extraction of the knowledge encoded within these models via natural language descriptions. Despite being in its infancy, this approach exhibits potentials to surpass traditional search engines as a superior source for knowledge and information acquisition.

Akin to refining queries for search engines, prompts given to LPMs must also be carefully crafted. However, the complexity of prompts, the unpredictability of model responses compared to traditional search engines present unique challenges. Significant research efforts \citep{jagerman2023query,zhong2023adapter} have been made to comprehend how LPMs react to various prompts, with some studies examining the feasibility of rewriting prompts for specificity. However, without access to users' personal data and behavior, tailoring the prompt to meet the user's needs accurately remains challenging.

Our research addresses this issue by integrating user preference information into prompt rewriting. The primary obstacle in personalized query rewriting is the absence of a dataset containing text-to-image prompts with personalized information. To overcome this, we have assembled a large dataset encompassing over 300k text-to-image histories from 3,115 users. We rewrite user prompts using their query history, although we had limited access to personal information, leaving room for further research. Another significant challenge is the evaluation of rewritten queries. To evaluate their efficacy, we've developed a new offline method that uses multiple metrics to measure how well our rewriting models can recover the original user query from the ChatGPT-shortened version.


Our paper's contributions are threefold:

1. We've compiled a large Personalized Image Prompt (PIP) dataset, which will be made public upon paper acceptance to aid future research in this field.

2. We experimented with two query rewriting techniques and proposed a new query evaluation method to assess their performance.

3. We propose a new benchmark for personalized text-to-image generation, which promotes the standardization of this field.

While there is still a considerable distance to cover before we can create a perfect prompt encapsulating both the user's requirements and the model's capabilities, we believe our research provides a critical stepping stone in this ongoing exploration.

\section{Related Work}
\label{sec:rw}

This section provides an overview of prior work on text-to-image generation, personalization for such generation, and prompt rewriting. It's important to note that our review is aimed more at offering sufficient background knowledge rather than exhaustive coverage of all related works.
\subsection{Text-to-Image Generation}
Large text-to-image generation models can generate high-fidelity image synthesis and achieve a deep level of language understanding.
DALL-E~\citep{ramesh2021zero} uses a VQ-VAE transformer-based method to learn a visual codebook in the first stage and then trains autoregressive transformers on sequences of text tokens followed by image tokens in the second stage.
DALL-E2~\citep{ramesh2022hierarchical} introduces latent diffusion models to generate various images by conditioning on CLIP text latents and CLIP image embeddings generated by a prior model.
Imagen~\citep{saharia2022photorealistic} discovers that a larger language model with more parameters trained on text-only data improves the quality of text-to-image generation. 
Late developments like stable-diffusion (SD)~\citep{Rombach_2022_CVPR} proposes to generate images effectively in latent space significantly lowering computational costs. Furthermore, SD designs a conditional mechanism to complete class-conditional, text-to-image and layout-to-image models. Furthermore, ControlNet~\citep{zhang2023adding} accomplishes certain function by conditioning on multi-modal data, e.g., edge, sketching, pose, segmentation, depth etc., which unavoidably involving additional condition-generation modalities.

Despite these models' ability to generate high-fidelity images, they often fail to meet the precise needs of the users. Text-to-image generation is more like a game of chance. 
\subsection{Personalization for Text-to-Image Generation}
Recently, personalization approaches based on text-to-image models have taken a set of images of a concept and generated variations of the concept. Specifically, some methods optimize a set of text embeddings. For example, ~\cite{cohen2022my} involves pseudo-word embeddings by a set encoder to provide personalization and Textual inversion~\citep{gal2022image} composes the concept into language sentences and performed as a personalized creation. 
Some methods finetune the diffusion model. For example, DreamBooth~\citep{ruiz2023dreambooth} finetunes the text-to-image diffusion model with shared parallel branches. 
To speed up, CustomDiffusion~\citep{kumari2023multi} reduces the amount of fientuned parameters, and ~\citet{tewel2023key} locks the subject's cross-attention key to its superordinate category to align with visual concepts. Moreover,  an additional encoder is trained to map concept images to its textual representation by ~\cite{gal2023encoder} and ~\cite{shi2023instantbooth}.

Existing studies have three key limitations: they demand extra images and fine-tuning of text-to-image models with limited scope for new concepts; they can't learn from user interaction history and need detailed user prompts; and there's a lack of public, personalized text-to-image datasets that truly reflect user preferences.

\subsection{Prompt Rewriting}
Recently, researchers have found that optimizing prompts can boost the performance of LLMs on several NLP tasks and even search systems. For examples, 
~\citet{guo2023connecting} connect the LLM with evolutionary algorithms to generate an optimized prompt from parent prompts, without any gradient calculation. 
~\citet{yang2023large} propose to use LLM as an optimizer by generating new prompts based on a trajectory of previously generated prompts in each optimization step with the objective of maximizing the accuracy of the task.
In the work~\citep{zhou2022large}, LLMs serve as models for engineering work like inference, scoring, and resampling. 
In search systems, LLMs are used to generate query expansion terms by ~\cite{jagerman2023query}, while they are used to reformulate query by ~\citet{wang2023generative} instead.

For T2I generation, one relatively close work, SUR-adapter~\citep{zhong2023adapter} learns to align the representations between simple prompt and complex prompt by an adapter. \cite{hao2022optimizing} optimize prompt through general rewriting. However, Neither of above works utilized personalized information for improving prompts.

\section{Personalized Image-Prompt (PIP) Dataset }

\subsection{Dataset Collection}
The Personalized Image-Prompt (PIP) dataset is the first large-scale personalized generated image-text dataset. 
The original data are collected from a public {website\footnote{https://zmrj.art/}} that we host to provide open-domain text-to-image generation to users. 
{PIP dataset includes 3115 users and 300,237 text-to-image histories generated by these users using SD v1-5~\citep{Rombach_2022_CVPR} and an internal fine-tuned version of SD v1-5. Each user in PIP has created 18 or more images and provided at least 12 different prompts.}


\begin{figure}[!t]
\begin{center}
\includegraphics[width=0.98\columnwidth]{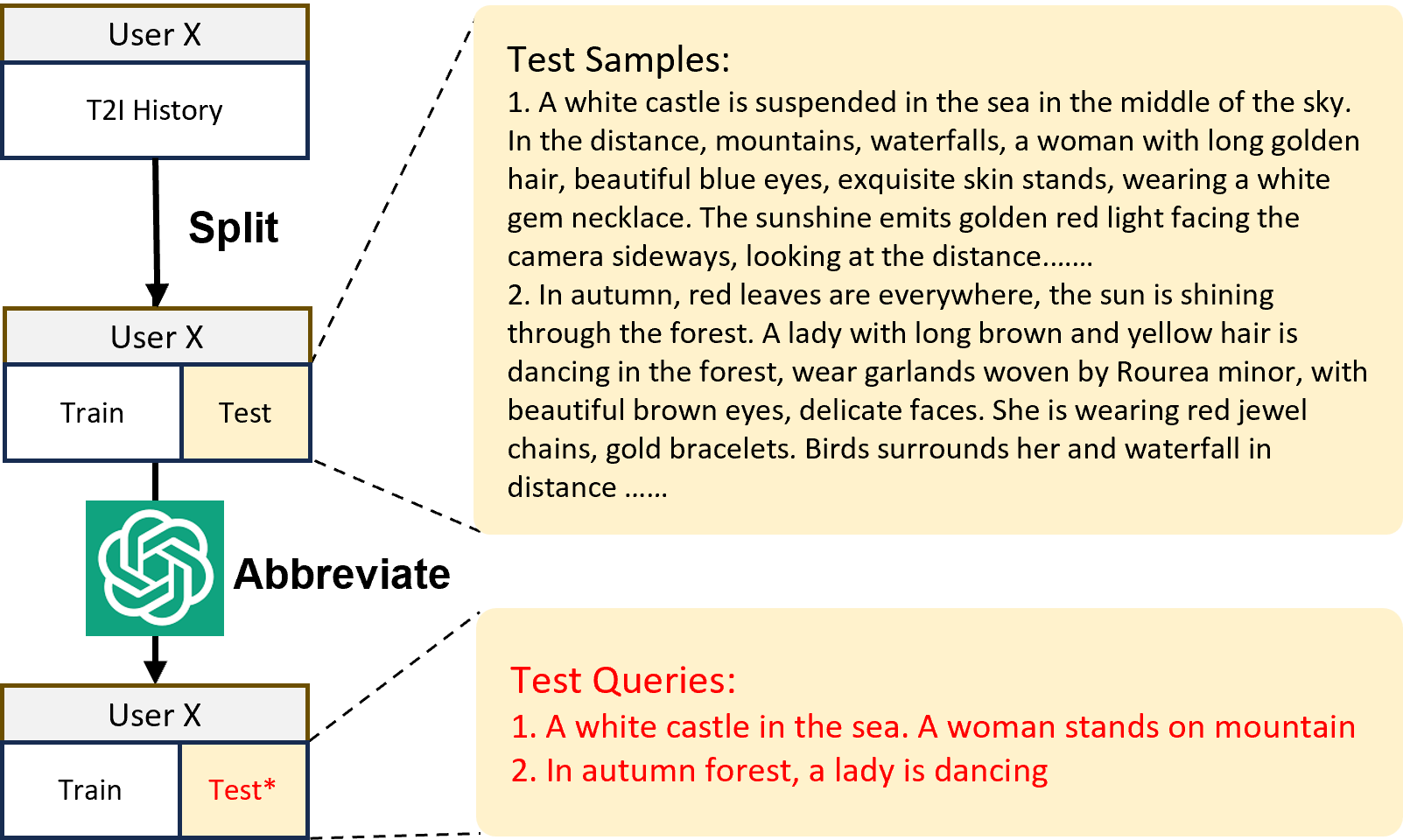}
\end{center}
\caption{\small Dataset creation process. We split our dataset into training and testing sets and summarize each prompts in the test set using ChatGPT. }
\label{fig:data_processing}
\end{figure}

\begin{figure}[!t]
\begin{center}
\includegraphics[width=\columnwidth]{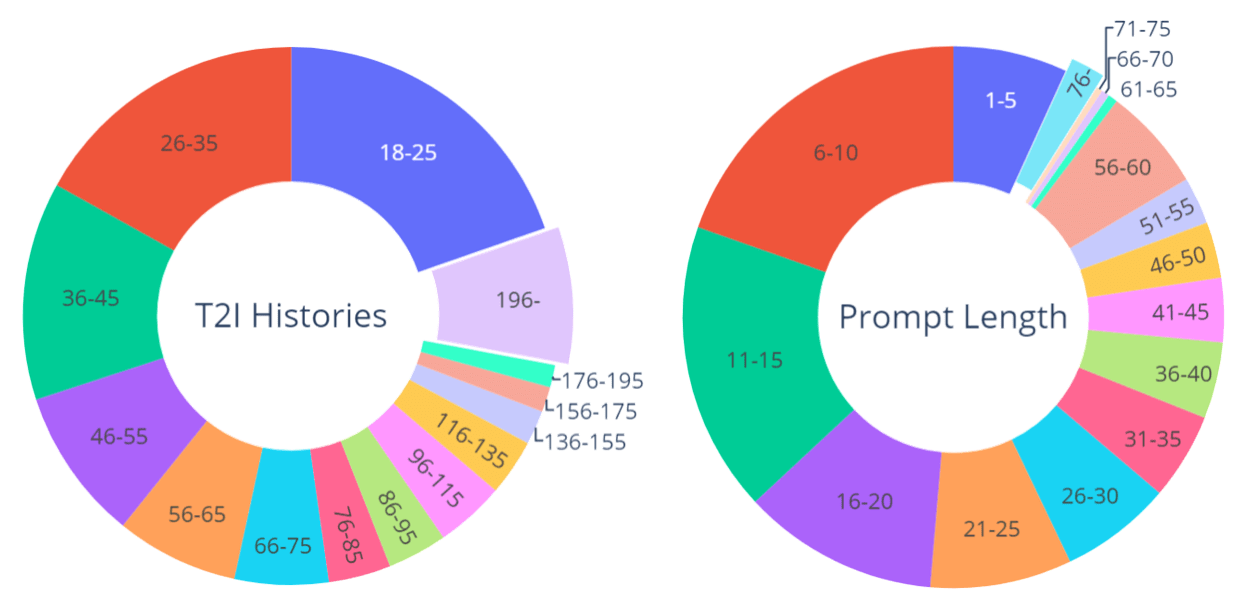}
\end{center}
\caption{\small Dataset statistics and distribution. \textbf{Left:} Proportion of users based on the varying number of historical prompts they have. Note that each user has a minimum of 18 historical prompts, as we have excluded those with fewer prompts from the dataset. \textbf{Right:} Proportion of prompts based on their varying lengths. Best view in color.}
\label{fig:data_stats}
\end{figure}

Figure~\ref{fig:data_processing} illustrates the process of creating the dataset. For each individual user, we randomly choose two prompts to serve as test prompts, with the remaining prompts allocated as training prompts (historical user query). The purpose of using random selection instead of the most recent generated prompts is to enhance the diversity of our test data. Subsequently, we employ ChatGPT to condense the test prompts, ensuring they only include the primary object or scene, as depicted in Figure ~\ref{fig:data_processing}. We shorten the prompts into three scales, i.e., contain only nouns, noun phases or short sentences respectively. 

In the ensuing experiment, each test prompt in the test set will be considered as the input prompt $x_t$ for every user $u$, with the original prompts serving as the ground truth that reflect the user's authentic preferences. The remaining prompts are utilized as training samples.

The PIP dataset consists of 300,237 image-prompt pairs, personally categorized by 3,115 users. These pairs are divided into 294,007 training samples and 6,230 test samples.

\subsection{Dataset Statistics and Distribution}
In this section, we showcase the data statistics that depict the quality and diversity of PIP. We specifically illustrate data distributions of the number of prompts and prompt length for each user, and delve deeper into the content of the prompt through a word cloud representation. 

\begin{figure}[htbp]
\centering
\includegraphics[width=0.98\columnwidth]{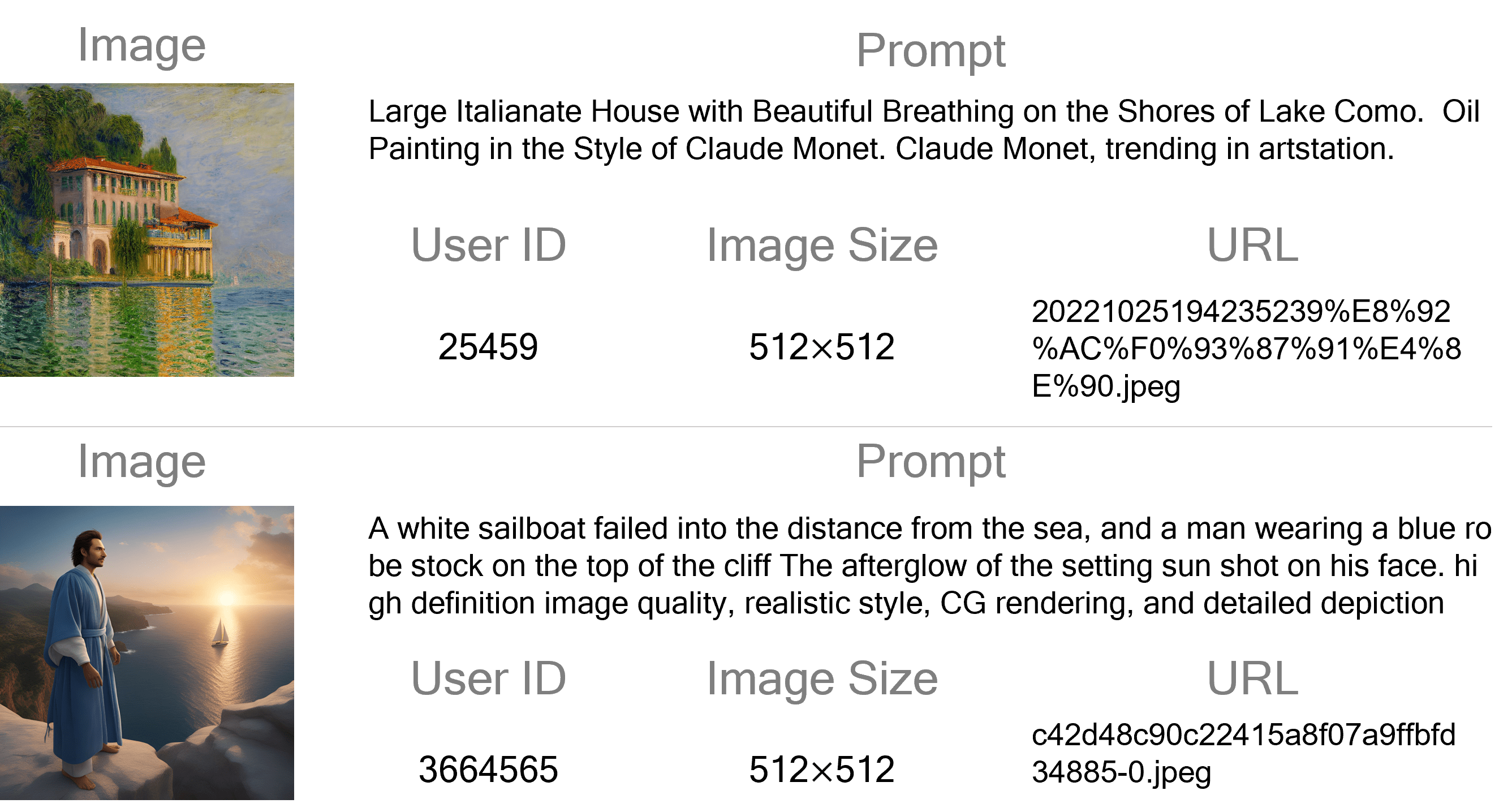}
\caption{\small Two examples from a user history, containing \texttt{Image}, \texttt{Prompt}, \texttt{User ID}, \texttt{Image size} and \texttt{URL}.}
\label{fig:data_exa}
\end{figure}

Each data sample contain a prompt, the generated images, UserID, Image size, and URL, as illustrated Figure~\ref{fig:data_exa}.

\begin{figure}[htbp]
\centering
\includegraphics[width=0.98\columnwidth]{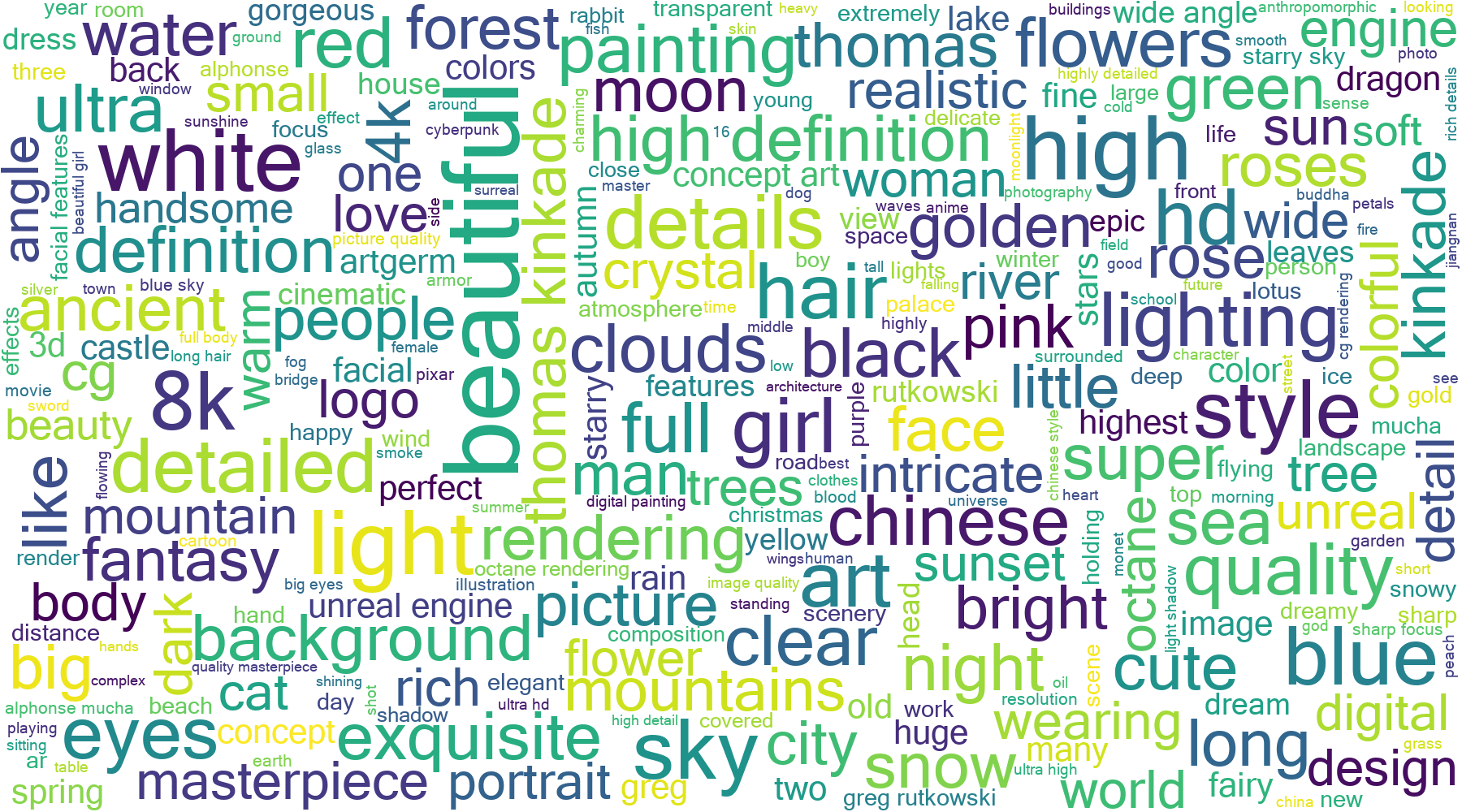}
\caption{\small Word cloud visualization of top 250 keywords sampled from the PIP dataset.}
\label{fig:exa_cloud}
\vspace{-2mm}
\end{figure}

In PIP dataset, each user contributes at least 18 images, as depicted in the left section of Figure~\ref{fig:data_stats}. This results in a long-tail distribution. The prompts have an average word count of 27.53. The length of the prompts, ranging from 1 to 284 words, also follows a long-tail distribution, as seen in the right section of Figure~\ref{fig:data_stats}. Despite the presence of about 2500 prompts that exceed the 75-word limit of SD, we retain them to maintain the integrity of user preferences.

Figure~\ref{fig:exa_cloud} presents the 250 most frequently used words or phrases. The frequency of these words is determined by the highest TF-IDF value across all users. The word cloud reveals that these words describe various image attributes, such as objects, styles, quality, and colors. This variety underlines the high diversity present within the prompt content of the PIP dataset.

\subsection{User Preference Evaluation}
\label{UserPref}
\begin{figure}[!t]
\centering
\includegraphics[width=\columnwidth]{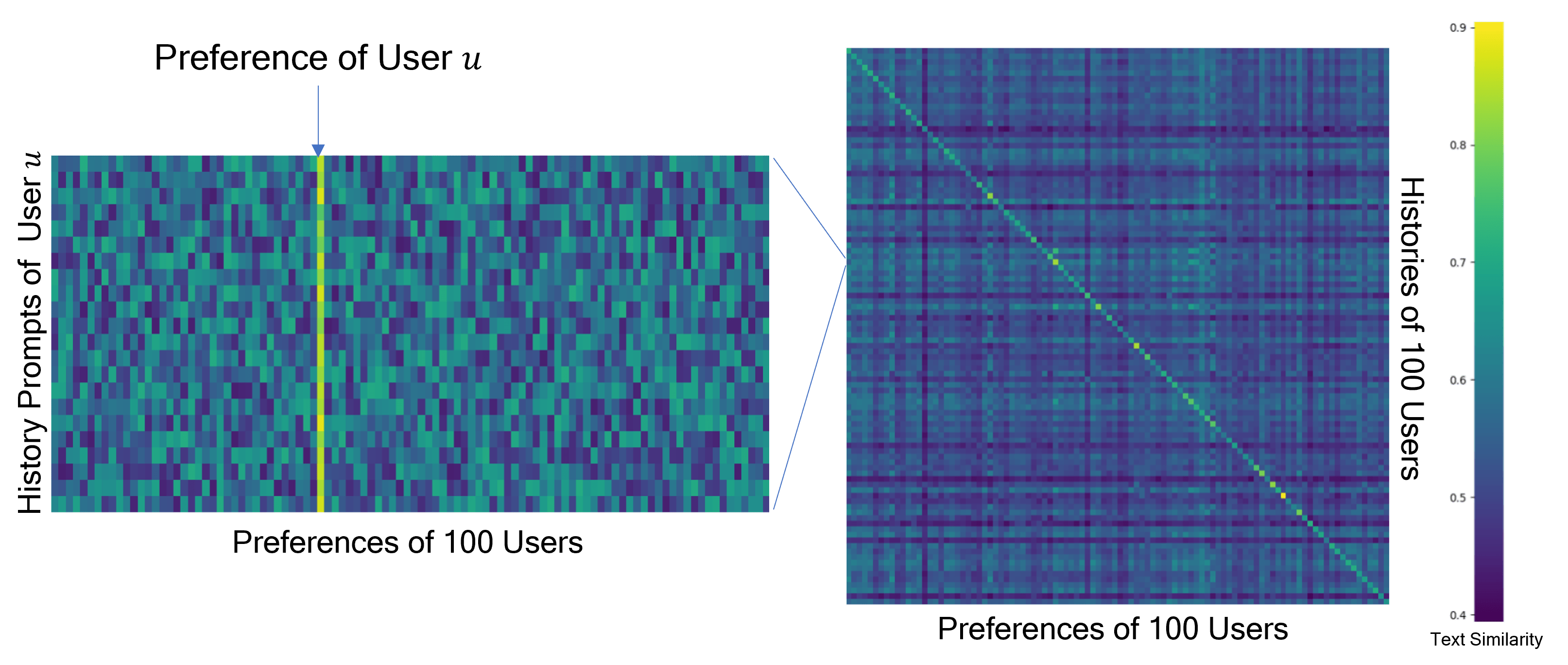}
\caption{Text similarity between user histories and user preferences. There is a great deal of diversity in user preferences in the PIP dataset. In the zoomed-in version of the similarity map on the left, users' history prompts are highly similar to their preferences. }
\label{fig:data_pref}
\end{figure}

For each user $u$ in PIP dataset, we summarize his preference $P_u$ into 5 phrases from his history prompts using ChatGPT. 

We visualize the text similarities between the histories and preferences of 100 users in 
Figure~\ref{fig:data_pref}. For two users $u,v$, the similarity of $u$'s histories and $v$'s preferences is defined as the mean of text similarities between $u$'s history prompts and $v$'s preference $P_v$. The text similarity is calculated using GTR-T5-large ~\cite{ni2021large}. Furthermore, we visualize the text similarities between the history prompts of a random user and 100 user preferences in the left part of Figure~\ref{fig:data_pref}. This shows that text-to-image users have different preferences and the preference $p_u$ we summarized successfully captures the key feature the user prefers.

Based on this observation, we present two metrics to evaluate prompt rewriting methods in terms of how the rewritten results are aligned to a user's preference, namely Preference Matching Score (PMS) and Image-Align.


\textbf{Preference Matching Score (PMS).} PMS calculates the CLIPScore~\citep{hessel2021clipscore} between generated image and user's preference $P_u$. 
It measures how the generated image aligns with the user's preference. 

\vspace{-5mm}
\begin{align}
\vspace{-4mm}
\textbf{PMS}=  \frac{w}{N} \sum^N_{u=1}\max\left(\text{cos}\left(\mathbf{Em}\left(I'_{u}\right),\mathbf{Em}\left(P_u\right)\right),0\right)
\end{align}
\noindent where $P_u$ is the user $u$'s preference, $I'_{u}$ is the generated image correspondingly, $N$ is total user number (i.e. 3115). $\mathbf{Em}$ means the embedding extracted by using CLIP. $w=2.5$ is a scaling constant.

\textbf{Image-Align.} It measures the similarity between the generated image and the ground-truth image. Image-Align quantifies how closely the current created image aligns with the user's truly saved image. The similarity between two images are calculated using CLIP ~\cite{radford2021learning}.

Apart from these metrics, we also adopt ROUGE-L to evaluate prompt rewriting methods in our experiment. Calculating ROUGE-L between the rewritten prompt against the original prompt measures the ability of prompt rewriting methods to recover the original prompt. We set $\beta=5$ to emphasize the recall of generated prompts.


\section{Personalized Prompt Rewriting}
\label{PromptRew}
\begin{figure*}[t]
\begin{center}
\includegraphics[width=\linewidth]{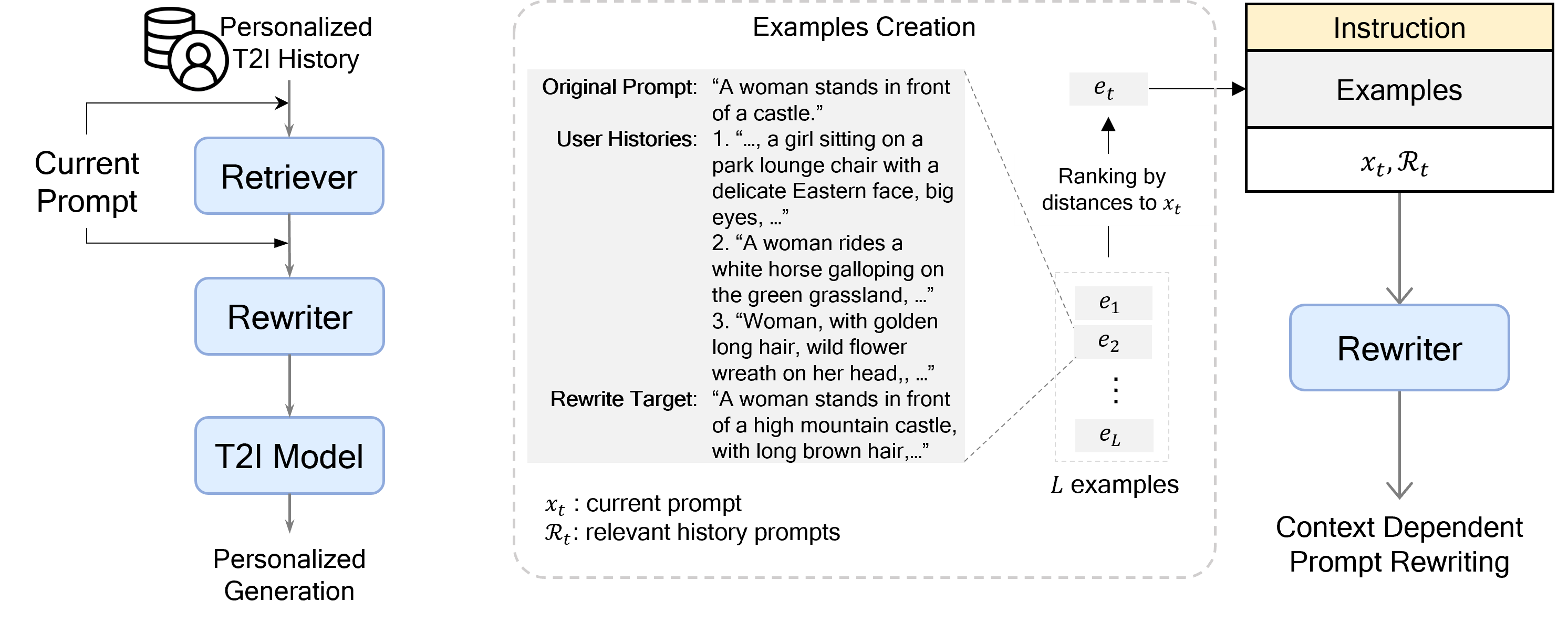}
\end{center}
\vspace{-3mm}
\caption{\textbf{Left:} Pipeline of Personalized Prompt Rewriting (Personalized PR), including Retriever, Rewriter and T2I model to generate personalized images from user histories. \textbf{Right:} Illustration of context-dependent prompt rewriting. We present a specific example for better understanding the procedure of context-dependent prompt rewriting.}
\vspace{-3mm}
\label{fig:Pipe}
\end{figure*}

The basic idea of our personalization method is to rewrite the input prompt, considering user preferences gleaned from past user interactions.
The full pipeline of our Personalized Prompt Rewriting (Personalized PR) method is depicted in the left part of Figure~\ref{fig:Pipe}.
If a user $u$ input a prompt $x_t$, a retriever $\mathbf{Ret}\left(x_t,\mathcal{Q}_t\right)$ retrieves prompts from the user's historical prompt set $\mathcal{Q}_t$, using $x_t$ as a query. 
Based on the retrival result $\mathcal{R}_t=\mathbf{Ret}\left(x_t,\mathcal{Q}_t\right)$, the rewriter $\mathbf{Rew}$ rewrites the input prompt to generate a personalized prompt $x'_t=\mathbf{Rew}\left(x_t,\mathcal{R}_t\right)$.
Finally, the text-to-image generation model $\mathbf{G}$ produces the image $I'_{t}=\mathbf{G}\left(x'_t,\epsilon\right)$ from the rewritten prompt, where $\epsilon$ is a random noise vector\footnote{The image generated from the prompt $x_t$ is denoted as $I_t$, where $I_t=\mathbf{G}\left(x_t, \epsilon\right)$.}.

\subsection{Retrieval and Ranking}
In the retrieval stage, given the input prompt $x_t$, the retriever $\mathbf{Ret}\left(x_t,\mathcal{Q}_t\right)$ retrieves relevant prompts from historical prompt set $\mathcal{Q}_t$, using $x_t$ as a query.

By analyzing user prompts, we have noticed that users tend to construct prompts that involve objects, their attributes, and the relationships between objects. 
In the works~\citep{schuster2015generating},~\citep{krishna2017visual}, a query for image retrieval is defined to include objects, attributes of objects, and relations between objects. 
Inspired by this, we suspect users have the habit of using attributes and some objects, such as background, to express their preferences. To confirm this, we visualize the word cloud of the top 250 frequent words in the text prompts of all users, as shown in the right part of Figure~\ref{fig:exa_cloud}.
In Figure~\ref{fig:exa_cloud}, we find some attributes, such as ``cute", ``golden", and ``beautiful" appear in prompts with high frequency, as well as some objects, such as ``mountain", ``sea," and ``sky".
Intuitively, we can use the current prompt $x_t$ to locate the relevant history prompts that include the same or similar attributes or objects.

To locate relevant prompts, two retrieval methods are used: dense and sparse.
In dense retrieval, we choose the prompt $x_t$ and calculate its textual embedding $\mathbf{Em}\left(x_t\right)$ using CLIP's text encoder, also the text encoder in Stable Diffusion~\citep{Rombach_2022_CVPR}. 
We suspect the prompts with similar visual attributes and objects will be close to each other in the text embedding space.
To confirm this, we visualize some retrieval results in Figure~\ref{fig:qualitative_analysis}.
The three nearest neighbors of $\mathbf{Em}\left(x_t\right)$ are prompts that are semantically related.
For example, if the input prompt is ``Hobbit homes", the three most relevant prompts would include the words ``village", ``city", and ``house".
This dense retrieval method is also referred to as embedding-based retrieval (EBR).
In sparse retrieval, we use BM25 to locate relevant prompts that include the same visual attributes and objects.

In the above retrival, we rank relevant prompts in EBR-based or BM25-based ranking, depending on the retrieval ways.
In EBR-based ranking, we rank the relevant prompts based on their embedding similarity with the query $x_t$. 
For similarity measuring, we choose cosine similarity as it is a commonly used similarity measure in embedding learning.
In BM25-based ranking, BM25 scores are used for similarity measures.
Consequently, we obtain the top $k$ relevant user queries $\mathcal{R}_t=\left\{r_1,...,r_k\right\}$.

\vspace{-3mm}
\subsection{Rewriting}
\label{Rewrite_32}
\vspace{-3mm}
The procedure of context-independent rewriting leverages pertinent queries $\mathcal{R}_t=\left\{r_1,...,r_k\right\}$, and employs ChatGPT to encapsulate user preferences and rewrite the prompt directly. These queries $\mathcal{R}_t$ are organized based on their relevance to $x_t$. 

In the context-dependent scenario, we initially create a collection of demonstration examples $\mathcal{E}=\left\{e_1, ..., e_L\right\}$ using manual design. We then select a small subset of these examples to serve as demonstrations for each rewriting task. Given the issue of order sensitivity in in-context learning, as highlighted in the study~\citep{liu2021makes}, we arrange the demonstration examples in a descending sequence based on their proximity to the input prompt $x_t$. The in-context rewriting process we employ is illustrated in the right section of Figure ~\ref{fig:Pipe}.

\section{Experiment}
We carried out experiments for prompt rewriting methods on our PIP dataset. We validate our method through both offline and online evaluation. And we further analyze the number of historical prompts for best extracting the users' preference by ablating top retrieval.
For offline evaluation, we use the aforementioned three metrics: PMS, Image-Align, and ROUGE-L.

For online evaluation, we carried out single blind experiment for recently active users on our website. Real user feedback is collected to evaluate our method. 

\begin{figure*}[!t]
\centering
\includegraphics[width=1.0\textwidth]{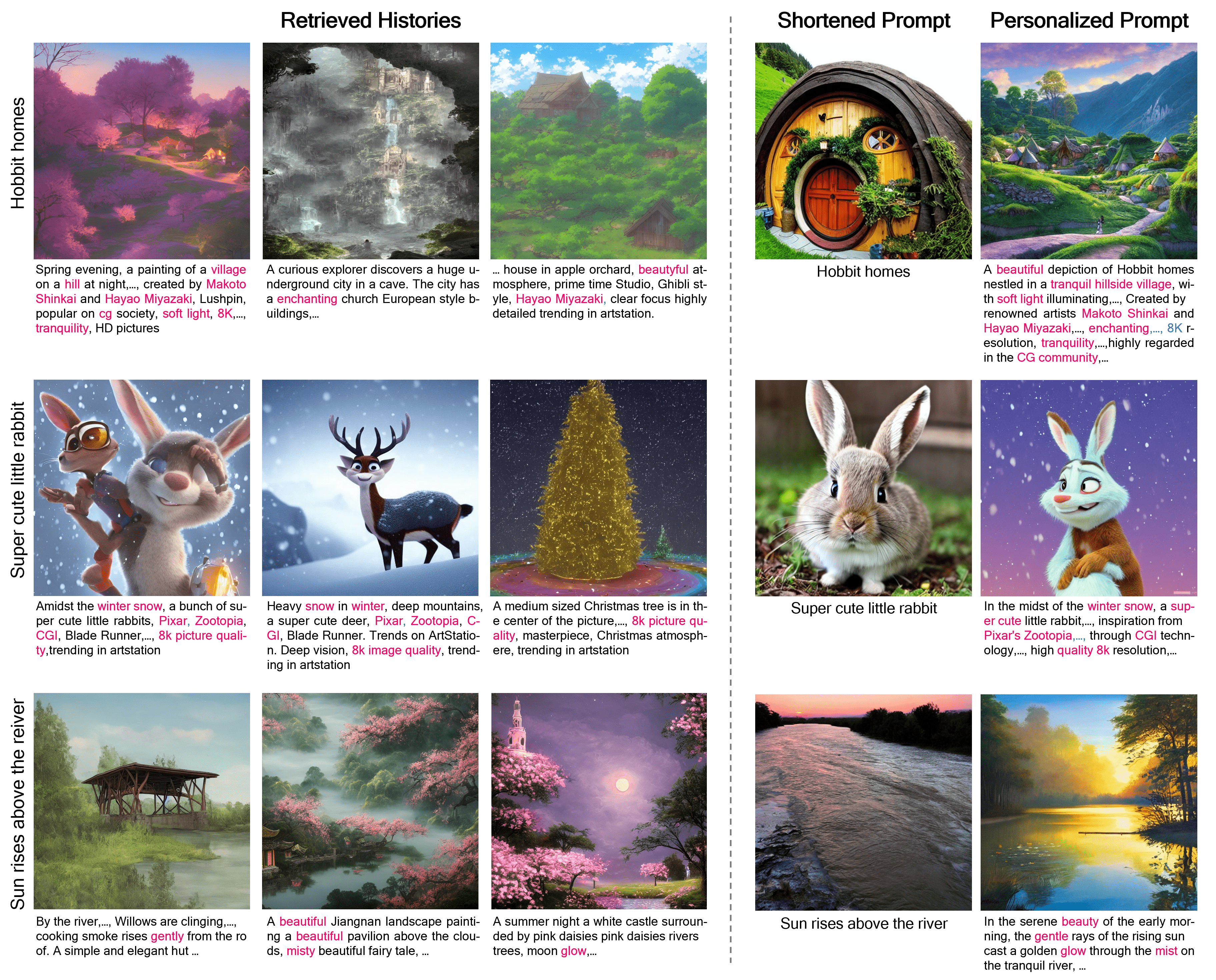}
\caption{Qualitative analysis of personalized retrieval and rewriting.}
\label{fig:qualitative_analysis}
\end{figure*}

\subsection{Implementation Details}
\label{imp_det}
\vspace{-1mm}

Details of our Personalized PR method are as follows. In retrieval, we choose the relevant text prompt number as $k=3$. 
We use ChatGPT~\cite{ChatGPT} as our rewriter. An input example for context-independent rewriting\footnote{The input for in-context rewriting is only different from context-independent rewriting in terms of the presence of demonstration example. See supplementary for detail.} is shown in Table ~\ref{tab:appendix prompt}. For in-context rewriting, we set $L=5$ and randomly select one demonstration example for each rewriting task, unless otherwise specified. 

Unless other specified, all the experiments use EBR to retrieve historical prompts, and one-shot in-context learning to rewrite the shortened prompt.

We use Stable Diffusion (SD) v1-5~\citep{SD_v15} as our text-to-image generation model for all methods.  SD v1-5 is sampled using PNDM scheduler in 50 steps and setting the classifier-free guidance scale to 7.0.

\begin{table}[!b]
    \centering
    \vspace{-3mm}
    \resizebox{1\columnwidth}{!}{
    \begin{tabular}{p{\columnwidth}}
    \toprule
    \textbf{\small ChatGPT Input Template for Context-independent Rewriting}\\
    \midrule
    \small Prompt in text-to-image generation describes the detailed attributes of the object user plans to draw. User's preference in text-to-image generation is shown in history prompts. \\
    \small Given 3 history prompts, your task is to rewrite the current prompt so that it matches the user’s preference. The rewritten prompt should retain primary objects in the original prompt and conform to the user’s preference. Please avoid being too diffused and restrict your output within 70 words.\\
    \small The history prompts are: \{$\mathcal{R}_t$\} \\
    \small The current prompt is: \{$x_t$\} \\
    \small The rewritten prompt (one sentence less than 70 words) is:\\
    \bottomrule
    \end{tabular}}
    \caption{Input template for context-independent rewriting.}
    \label{tab:appendix prompt}
\end{table}




We compare our method with two baseline methods, namely Promptist~\cite{hao2022optimizing} and General Prompt Rewriting (General PR). Both are general text-to-image prompt rewriting methods completely overlooking users' preferences. 

\textbf{Promptist~\cite{hao2022optimizing}}. Promptist uses the GPT-2 as the base model and has trained it on a self-collected text-to-image prompt dataset with 360K prompts, by using supervised fine-tuing method and reinforcement learning. 

\textbf{General PR}. We prompt ChatGPT~\citep{ChatGPT} to perform general prompt rewriting in a normal manner. In practice, we input ChatGPT with the current prompt $x_t$ without any user historical information.
Apart from the absence of histories, everything else of the prompt template for ChatGPT remains the same as the template in Table~\ref{tab:appendix prompt}. 

We feed baseline methods with the input prompt $x_t$ and obtain its output as prompt rewriting results $x_t'$. $x_t'$ is later used to generate their image results via $I'_{t}=\mathbf{G}\left(x'_t,\epsilon\right)$

\subsection{Qualitative Analysis}
To illustrate the effectiveness of our retrieval methods, we provide visualizations of the retrieved relevant image-prompt pairs in the left section of Figure~\ref{fig:qualitative_analysis}. These examples are drawn from the experimental results of three different test prompts from three users. The retrieved histories exhibit a high degree of similarity with their corresponding queries in terms of objects, attributes, as well as the overall style or mood of the image.

This effectively demonstrates the proficiency of our retriever in sourcing relevant user histories, thereby providing a robust reference for our rewriter to carry out personalized prompt rewriting.

In the right section of Figure~\ref{fig:qualitative_analysis}, we display the generation outcomes of both ``Shortened Prompt" and ``Personalized Prompt". The ``Shortened Prompt" column exhibits the results produced from the shortened prompts, while the ``Personalized Prompt" column features the rewritten prompts of our method along with their corresponding generated images. It's evident from these displays that images our method generates are more inclined towards user preferences based on their histories, a testament to the expressive power of our rewritten prompt. For instance, when the query ``Hobbit homes" is used (as seen in the first row), we observe the user's preferred style across three images, all capturing the mood of mountainous scenery and depicting the Hobbit homes within a consistent landscape.

\subsection{Quantitative Comparison}
\label{quantitative exp}

To further examine the effectiveness of our Personalized PR method, we conduct offline and online quantitative evaluation, comparing different settings to the baseline methods without enhanced prompt rewritten on the test samples we created. 

\begin{table}[h]
    \centering
    \resizebox{\columnwidth}{!}{
    \begin{tabular}{|c | c | c c c|}
    \hline
     Method & Retriever &  PMS $\uparrow$ & Image-Align$\uparrow$& ROUGE-L$\uparrow$  \\
    \hline
    \hline
    Shortened Prompt& -& 0.5567 & 0.6272 &0.3268  \\
    Promptist~\cite{hao2022optimizing} & -  & 0.5858 & 0.6481& 0.2947 \\
    General PR & -  & 0.5996&  0.5912& 0.2082\\
    \hline
    \hline
     \multirow{2}{*}{Personalized PR} & BM25 & 0.6125  &0.6581&0.3942\\
     & EBR   &0.6083 &0.6485 &0.4137 \\
    \hline
    \hline
     \multirow{2}{*}{Personalized PR+ ICL} & BM25  & \textbf{0.6253} &0.6456& 0.4417 \\
      & EBR   & 0.6179 &\textbf{0.6796} &\textbf{0.4686}\\
    \hline
       \end{tabular}}
       \vspace{-2mm}
    \caption{Comparison results of different variants of our method with the baseline. Evidently, our method using EBR retriever (top-3 retrieval) and 1-shot ICL can achieve most best results.}
    \vspace{-3mm}
    \label{tab:table 3}
\end{table}


\textbf{Offline Test} Table~\ref{tab:table 3} showcases the numerical results comparing various retrieval and rewriting configurations with shortened prompts.
We can see that Promptist~\cite{hao2022optimizing} and General PR are slightly better than `Shortened Prompt' in terms of PMS and Image-Align. 
Without the users' histories, these large language models perform prompt rewriting in arbitrary pathways. Thus, the rewritten prompts these baseline methods generate  could totally deviate from the users' preferences. Such hallucination scenarios result in the decrease in terms of ROUGE-L for both Promptist~\cite{hao2022optimizing} and General PR.
This indicates that although general prompt rewriting methods appear to be refining the prompts, their results do not align with users' preferences well. 
Our method outperforms all baseline methods, i.e. Promptist~\cite{hao2022optimizing} and General PR, on all metrics. 

A comparison between BM25 and EBR reveals that dense retrieval generally outperforms sparse retrieval, although the difference is relatively small, indicating that both methods can produce satisfactory results. 


Further, when contrasting context-independent rewriting with in-context rewriting, it's evident that ICL produces superior outcomes. By integrating ICL with EBR, we achieve absolute improvements of 14.2\%, 6.9\% and 5.6\% in terms of ROUGE-L, PMS and Text-Align metrics respectively. This underscores the exceptional performance of personalized prompting.

To check how sensitive our prompt rewriting methods with respect to the length of the prompts before rewriting. we assess our rewriting method using two additional types of prompts with shorter lengths, namely ``Noun Phrase" and ``Noun", as detailed in Table ~\ref{tab:table 2}.

\begin{table}[h]
    \centering
    \resizebox{\columnwidth}{!}{
    \begin{tabular}{|c | c | c c c|}
    \hline
    Prompt Type & Method & PMS $\uparrow$ & Image-Align$\uparrow$&ROUGE-L$\uparrow$\\
    \hline
    \hline
    \multirow{2}{*}{Noun} &Original&   0.5537 &0.6087 &0.1770 \\
    & Personalized PR & 0.6142 &0.6478&0.2804 \\
    \hline
    \multirow{2}{*}{Noun Phrase} & Original &  0.5554& 0.6146 & 0.2459\\
    & Personalized PR & 0.6168 &0.6534 & 0.3387\\
    \hline
    \multirow{2}{*}{Short Sentence} & Original &  0.5567 & 0.6272 & 0.3268\\
    & Personalized PR & 0.6179 & 0.6796 & 0.4686 \\
    \hline
    \end{tabular}}
    \caption{Performance with respect to different prompt lengths, i.e. \textit{Noun}, \textit{Noun Phrase} and \textit{Short Sentence}. Our ``Personalized PR" equipped with top-3 dense retriever, and 1-shot ICL consistently enhances results, even with only nouns or noun phrases.
 } 
    \label{tab:table 2}
\end{table}


These two prompts are derived using spaCy~\citep{Honnibal_spaCy_Industrial-strength_Natural_2020} from our dataset, adhering to the principle of minimizing word count while maintaining the main entities. The results displayed in Table ~\ref{tab:table 2} show that across all three shortened scale,  ``Personalized PR" outperforms the baseline across all metrics, demonstrating the effectiveness of our methods in recovering user preferences. 

\begin{figure}[!t]
    \centering   \includegraphics[width=\columnwidth]{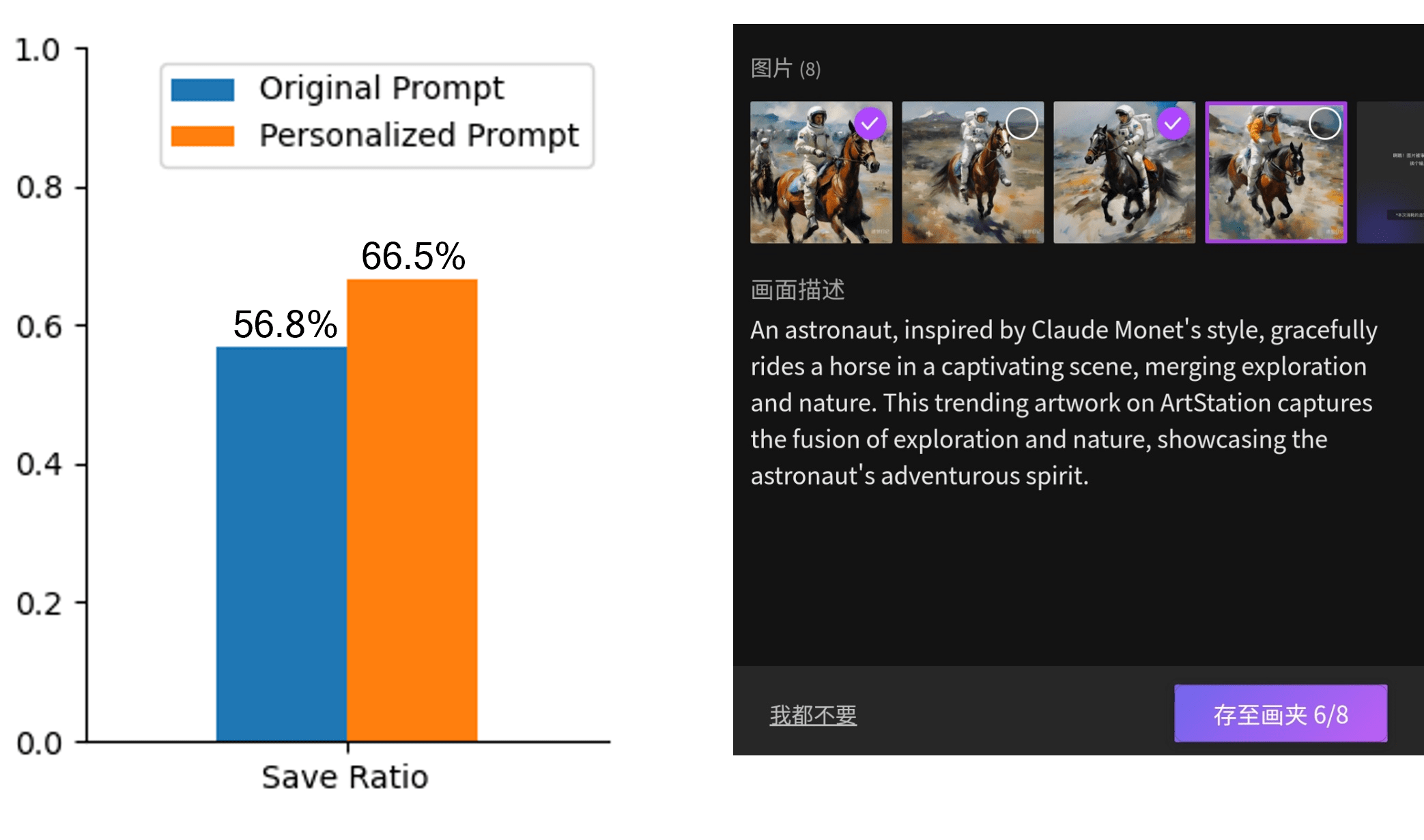}
    \caption{Illustration of our online evaluation methods and the improvement of Save Rate from personalized Prompt. } 
    \vspace{-5mm}
    \label{fig:online}
\end{figure}

\textbf{Online Test} To further validate our approach, we 
{have carried out a single-blinded online evaluation on our website mentioned above. Active users recently in the website are randomly selected to participate in the online test. Upon each prompt input by a participant, there's an equal chance of generating an image using either ``Original Prompt" or our method's ``Personalized Prompt". Participants can choose to ``Save" or ``Delete" each generated image based on their preference. We use their ``Save" actions to assess our method's effectiveness. In the online test, 247 users generated 905 images, with 433 from ``Original Prompt" and 472 from ``Personalized Prompt". The results of the online evaluation, as shown in Figure ~\ref{fig:online}, indicate that users prefer images generated using the ``Personalized Prompt" over the ``Original Prompt", with a 17.1\% increase in ``Save" actions.}
This suggests that our method aligns better with user preferences, affirming its effectiveness. We anticipate even better results in real-world scenarios with more user information and an improved text-to-image generation method.

\subsection{Ablation Study}
\label{ablat_stu}

In this section, we ablate $k$ relevant historical prompts used to rewrite personalized prompts and the number of demonstration examples used for in-context learning. We evaluate each experiment on the same setup as in Section ~\ref{imp_det}.

\begin{table}[h]
    \centering
    \resizebox{1\columnwidth}{!}{
    \begin{tabular}{|c | c | c c c|}
    \hline
     Method &Retrieval Top-$k$ &  PMS $\uparrow$&Image-Align $\uparrow$ & ROUGE-L $\uparrow$ \\
    \hline
    \hline
     \multirow{4}{*}{Personalized PR + ICL} & 1 & 0.6057 &0.6751 & 0.4539 \\
      & 3 & 0.6179 & \textbf{0.6796}& \textbf{0.4686} \\
     & 5 & 0.6204& 0.6748& 0.4474\\
     & 7 & \textbf{0.6265}&0.6651 & 0.4592 \\
    \hline
    \end{tabular}}
    \caption{Ablation for Retrieval Top-k. We empirically conduct experiments with respect to $k \in \{1,3,5,7\}$, keeping the configuration of DENSE retriever and 1-shot ICL.}
    \vspace{-3mm}
    \label{tab:abla-retrieval-k}
\end{table}


\textbf{Retrieval Top-k Ablation.} As shown in Table ~\ref{tab:abla-retrieval-k}, when using 3 most relevant historical prompts for rewriting, we can obtain most best results among all evaluation metrics. We analyze that too more historical prompts could provide redundant information and also too long prompt input to ChatGPT worsens the rewriting performance.
Therefore, we choose 3 as the number of retrieval results, a balanced manner between performance and efficiency.

\begin{table}[h]
    \centering
    \resizebox{1\columnwidth}{!}{
    \begin{tabular}{|c | c |c c c|}
    \hline
     Method (Retrieval) & ICL Shot  &  PMS $\uparrow$& Image-Align $\uparrow$& ROUGE-L $\uparrow$\\
    \hline
    \hline
     \multirow{3}{*}{Personalized PR (BM25)} & 1 & 0.6253 &0.6456& 0.4417 \\
     & 3 & 0.6289&0.6580& 0.4381\\
     & 5 & 0.6236& 0.6571& 0.4226 \\
    \hline
    \hline
     \multirow{3}{*}{Personalized PR (EBR)} & 1 & 0.6179 & \textbf{0.6796}& \textbf{0.4686}\\
     & 3 & \textbf{0.6274}& 0.6708 & 0.4354\\
     & 5 & 0.6242& 0.6724 & 0.4439\\
    \hline
    \end{tabular}}
    \caption{Ablation for Number of ICL Demonstrations. We experiment with consistent top-3 retrieval both on BM25 and EBR. When we set ICL shot as 1 and use EBR retriever, we observe more superior results appearing.} 
    \vspace{-3mm}
    \label{tab:abla-icl-l}
\end{table}


\textbf{Number of ICL Demonstrations Ablation.} Table ~\ref{tab:abla-icl-l} shows that 1-shot setting, i.e., given 1 demonstration example for in-context learning, can achieve the best results in general among all the four evaluation metrics regarding both BM25 and EBR. This demonstrates that our prompt rewriting template is efficient and effective enough for extracting the personalized preference from numerous historical data of each user.

\vspace{-1mm}
\section{Conclusion and Future Work}
\vspace{-1mm}
In conclusion, this study has underscored the significance of harnessing historical user behaviors to construct personalized AI content generation.
Our strategy aims to refine user prompts by capitalizing on previous user interactions with the system. We have introduced an innovative technique that entails reconfiguring user prompts based on a newly developed, large-scale text-to-image dataset encompassing over 300,000 prompts from 3,115 distinct users. This methodology has proven to augment the expressiveness of user prompts and ensure their alignment with the desired visual outputs. Our empirical results have underscored the supremacy of our techniques over conventional methods. This superiority was corroborated through our novel offline evaluation method and online tests, thereby affirming the efficacy of our approach.

Although the outcomes are encouraging, considerable research still lies ahead in this domain. In the realm of personalized text-to-image generation, the incorporation of more personal details like a user's age and gender could bolster performance. The methodologies used could be extended to other LPMs, including LLMs. Additionally, techniques for enhancing search engines, e.g. data source purification and ranking optimization, could be assimilated into these models. 

We are confident that our contributions represent advancements towards a more personalized and user-focused artificial intelligence.

{
    \small
    \bibliographystyle{ieeenat_fullname}
    \bibliography{main}
}

\clearpage
\setcounter{page}{1}
\maketitlesupplementary
\section{Details of User Preference Summarization in PIP Dataset}

Here, we provide details about how the user preference $P_u$ in PIP dataset is summarized. As mentioned in Section \ref{UserPref}, we use ChatGPT to summarize $P_u$ based on the user $u$'s historical prompts $\mathcal{Q}_t$. Considering ChatGPT's context length, we randomly select 50 historical prompts for summarization, assuming users have more than 50 different historical prompts. The input template for ChatGPT is shown in Table \ref{tab:pref_sum}. 

\begin{table}[h]
    \centering
    \begin{tabular}{p{\columnwidth}}
    \toprule
    \textbf{\small ChatGPT Input Template for Summarizing User Preference}\\
    \midrule
    \small Your task is to use no more than five phrases to summarize user's preference based on history text prompts he uses in text to image generation. \\
    \small A user's preference reveals what kind of image he might prefer or the image style he likes. Do not include objects that appear in history prompts in your answer. Just answer the phrases in sequential order, separate using a comma. \\
    \small Please summarize the preference for the following user: \\
    \small The history prompts of a user: $\mathcal{Q}_t$\\
    \small The keywords of the user's preference:
    \\
    \bottomrule
    \end{tabular}
    \caption{Prompt Template we use for summarizing user preference}
    \vspace{-3mm}
    \label{tab:pref_sum}
\end{table}

\section{Further Details of In-Context Personalized PR and General PR}
As mentioned in \ref{imp_det}, our prompt manner for in-context rewriting is as similar as for context-independent, both in Personalized PR and General PR. Here we show the in-context personalized PR input template in Table \ref{tab:ICL Rewrite}, and we show the general PR input template in Table \ref{tab:ChatGPT rewrite}, respectively.

\begin{table}[!t]
    \centering
    \begin{tabular}{p{\columnwidth}}
    \toprule
    \textbf{\small ChatGPT Input Template for In-context Rewriting}\\
    \midrule
    \small Prompt in text-to-image generation describes the detailed attributes of the object user plans to draw. User's preference in text-to-image generation is shown in history prompts. \\
    \small Given 3 history prompts, your task is to rewrite the current prompt so that it matches the user’s preference. The rewritten prompt should retain primary objects in the original prompt and conform to the user’s preference. Please avoid being too diffused and restrict your output within 70 words.\\
    \small Examples:\{$\mathcal{E}$\} \\
    \small The history prompts are: \{$\mathcal{R}_t$\}\\
    \small The current prompt is: \{$x_t$\}\\
    \small The rewritten prompt (one sentence less than 70 words) is:\\
    \bottomrule
    \end{tabular}
    \caption{ChatGPT input template for in-context personalized prompt rewriting.}
    \label{tab:ICL Rewrite}
\end{table}

\begin{table}[!t]
    \centering
    \begin{tabular}{p{0.98\columnwidth}}
    \toprule
    \textbf{\small ChatGPT Input Template for General Prompt Rewriting}\\
    \midrule
    \small Prompt in text-to-image generation describes the detailed attributes of the object he plans to draw. A nice prompt for text-to-image generation usually includes various aspects of the image including descriptions of the scene, mood, style, lighting, and more. \\
    \small Given an input prompt, your task is to rewrite the prompt to a better one. Your rewritten prompt is supposed to describe the image better, and less than 70 words. \\
    \small The input prompt is: \{$x_t$\} \\
    \small The rewritten prompt (one sentence less than 70 words) is : \\
    \bottomrule
    \end{tabular}
    \vspace{-2mm}
    \caption{Input template for General PR.}
    \vspace{-3mm}
    \label{tab:ChatGPT rewrite}
\end{table}

\begin{figure*}[]
    \centering
    \includegraphics[width=\textwidth]{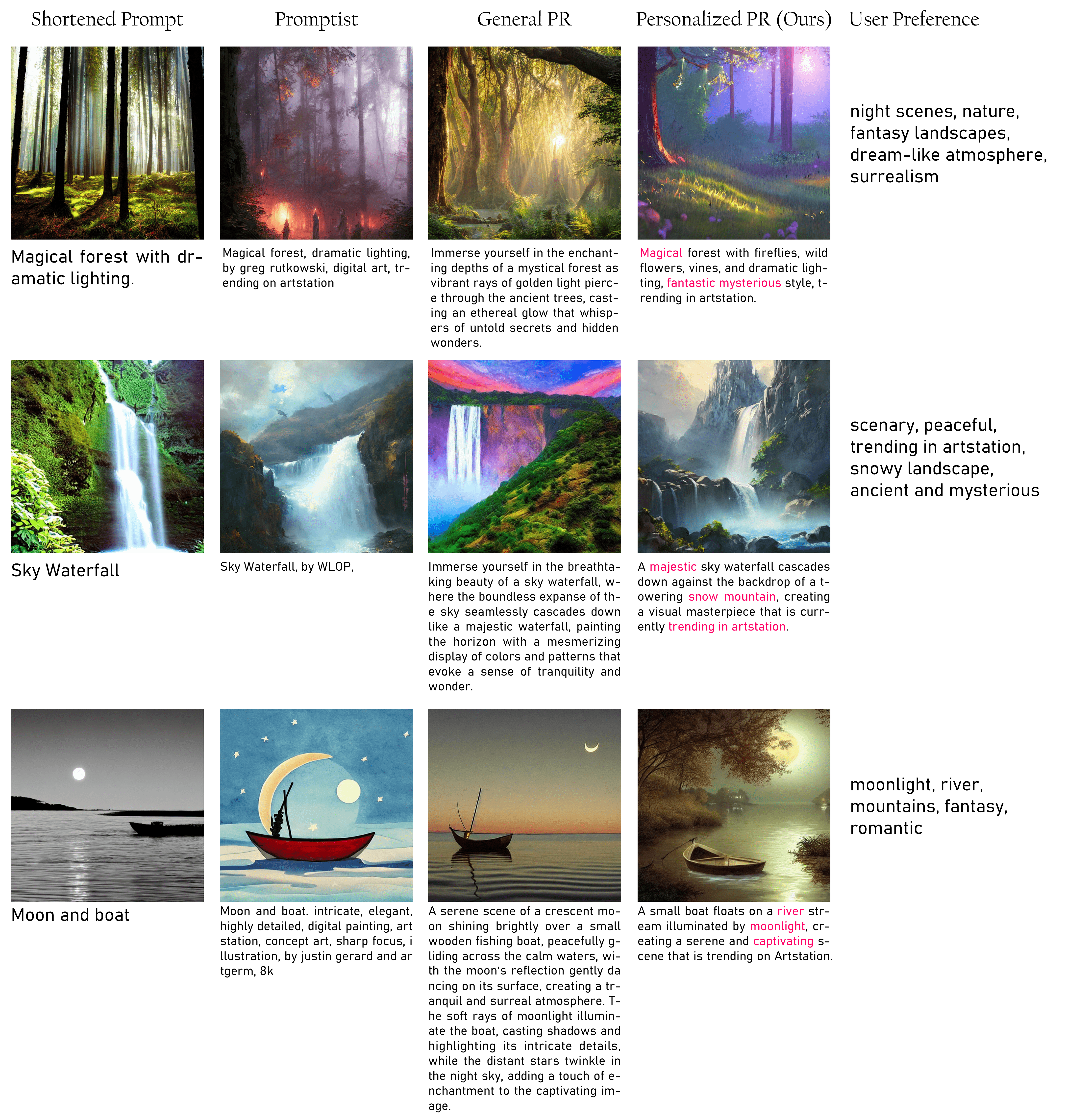}
    \caption{Qualitative comparison of our method Personalized PR against baseline methods. Compared with Promptist, General PR, our personalized method aligns with user preference better. Please zoom in to see details.}
    \label{fig: appen rs1}
\end{figure*}

\begin{figure*}[]
    \centering
    \includegraphics[width=\textwidth]{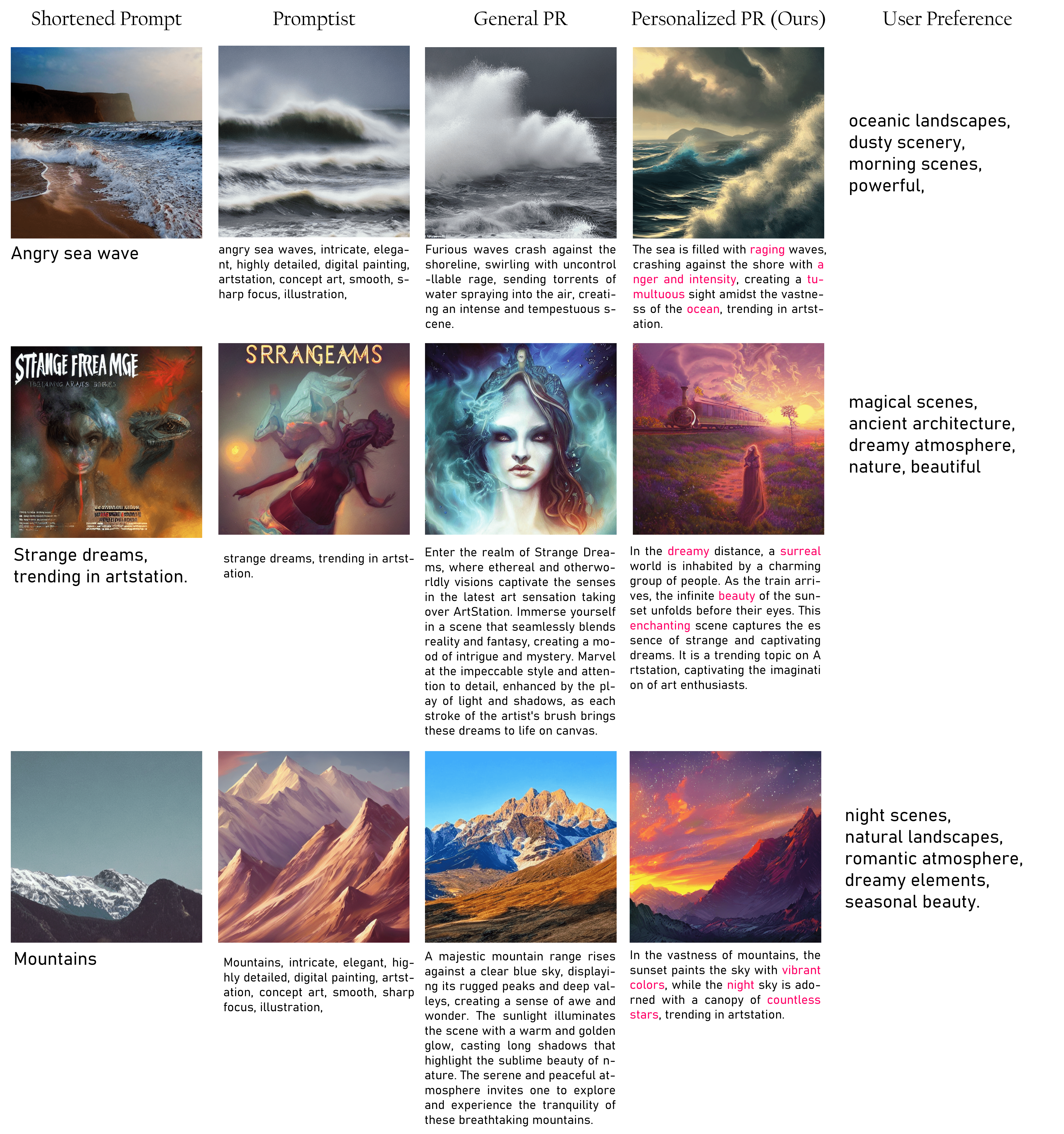}
    \caption{More comparison results. Please zoom in to see details.}
    \label{fig: appen rs2}
\end{figure*}

\begin{figure*}[]
    \centering
    \includegraphics[width=\textwidth]{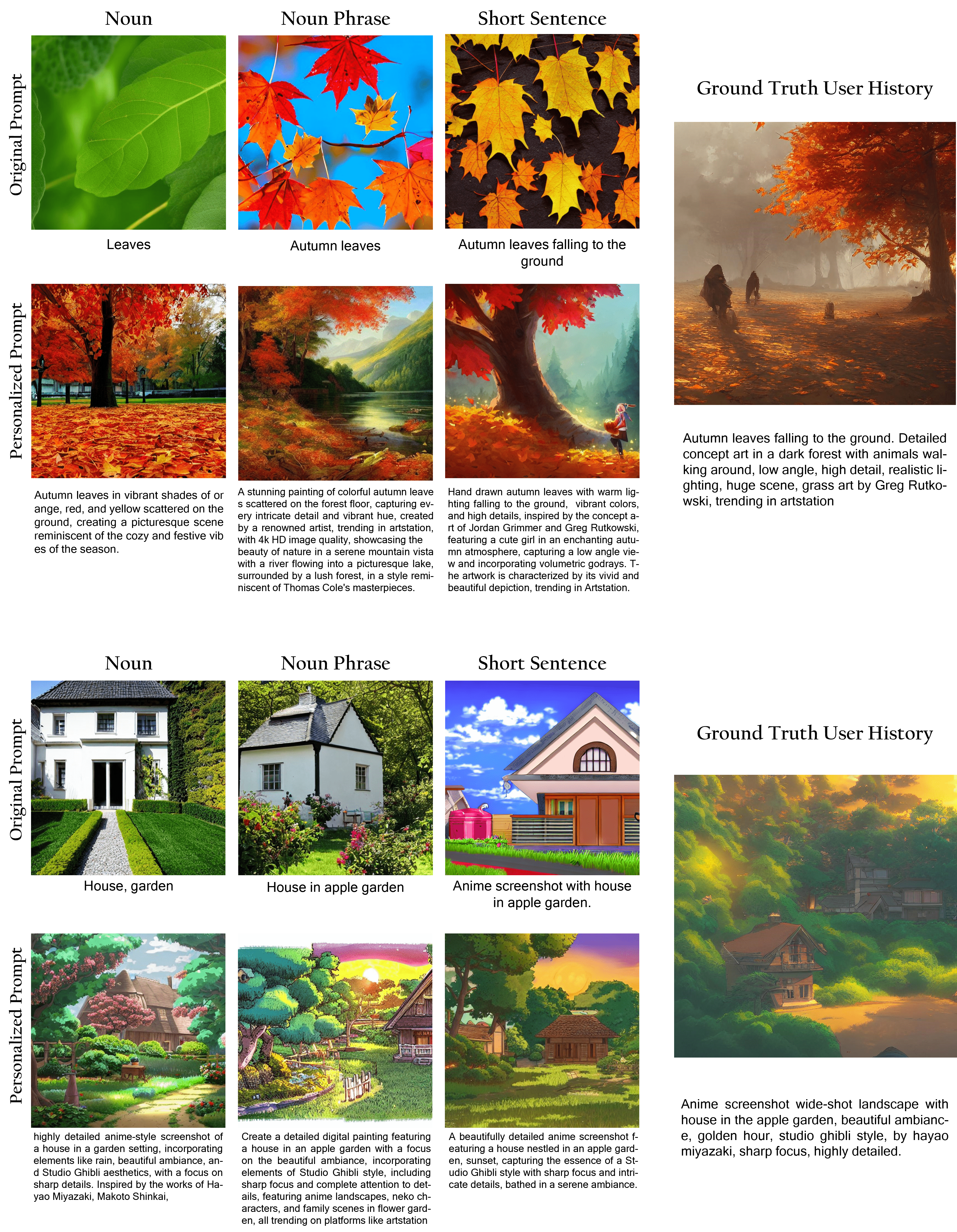}
    \caption{Qualitative results of our method under different type (length) of input prompts. Our personalized PR can generate more expressive user-preferred images as similar as the user groundtruth compared with the original prompt. Please zoom in to see details.}
    \label{fig: appen 3l}
\end{figure*}

\begin{figure*}[]
    \centering
    \includegraphics[width=\textwidth]{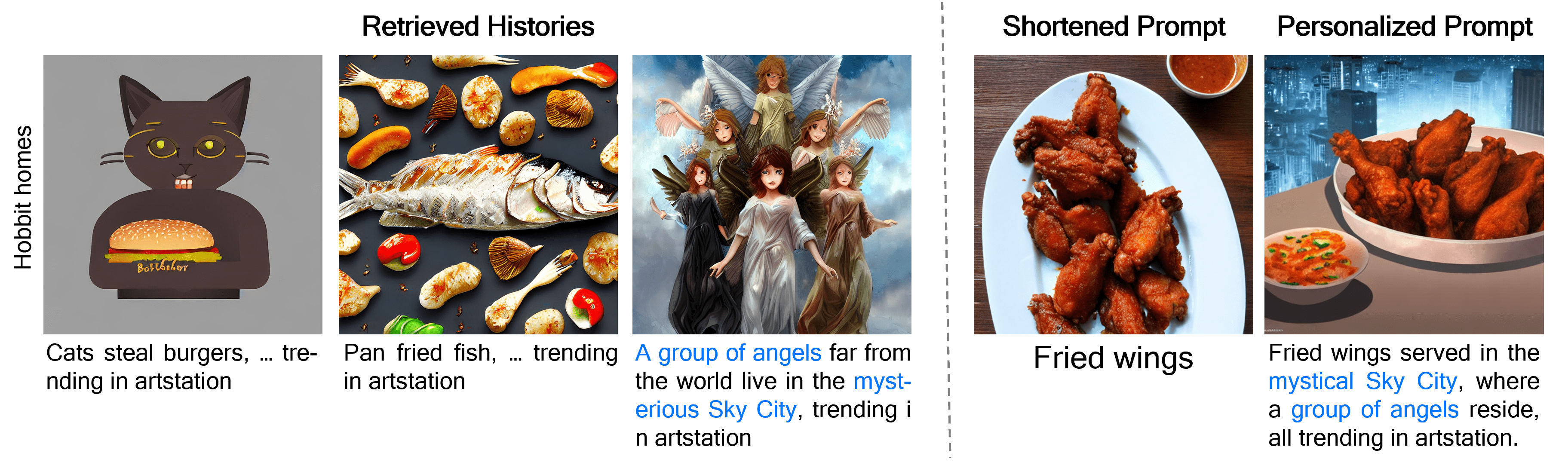}
    \caption{A failure case of our method. The words in \textcolor{blue}{blue} indicate irrelevant objects are added to the rewritten prompt, resulting bad generation.}
    \label{fig: appen fail1}
\end{figure*}

\begin{figure*}[]
    \centering
    \includegraphics[width=\textwidth]{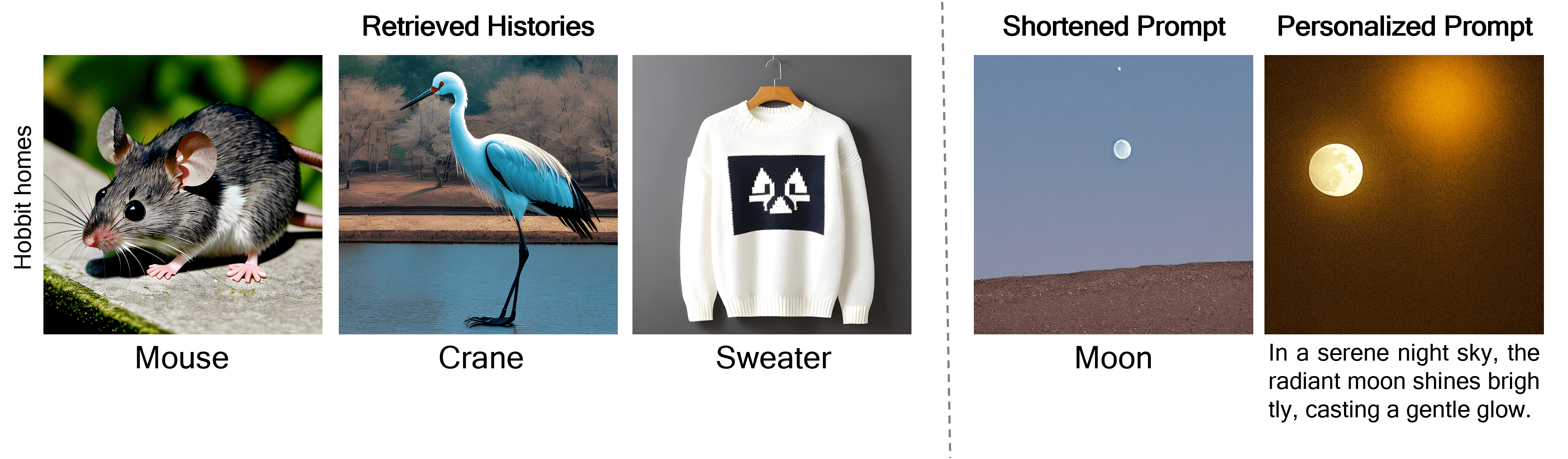}
    \caption{A failure case of our method.}
    \label{fig: appen fail2}
\end{figure*}

\section{Additional Qualitative Results}
We provide additional qualitative results of our Personalized PR method to demonstrate the effectiveness of our method.

In Figure \ref{fig: appen rs1} and Figure \ref{fig: appen rs2}, we show comparisons of our method against baseline methods, i.e. Promptist \cite{hao2022optimizing} and General PR, in terms of the rewritten prompt as well as the corresponding images generated using SD v1-5. All methods use the "Shortened Prompt" as input $x_t$. To evaluate the alignment with the user preference of the compared methods, we show the user preference $P_u$ for several cases of the exemplar. 
By leveraging historical prompts, our method can rewrite the prompt in accordance with the user's preference in each case. However, general prompt rewriting methods either insert just a few tags (such as ``digital art" or ``by greg rutkowski" etc.) arbitrarily. These generations either produce something resembling a carefully crafted text-to-image prompt or continually add tedious details, resulting in a lengthy text output. 
Lost histories, these methods fail to meet the user's preference.


As mentioned in \ref{quantitative exp} and Table \ref{tab:table 2}, we have conducted experiments to demonstrate the performance of our method in terms of two shorter types of input prompt $x_t$. The quantitative results in \ref{tab:table 2} showcase that our method performs robustly even on conditions of very short prompt. Here, we provide two qualitative examples along with user ground truth generated images in Figure ~\ref{fig: appen 3l}. These results showcase that, by leveraging historical prompts, our method is capable of rewriting input prompt $x_t$ properly and further generating images that are very close to the users' true intentions, either in terms of style or objects. 
Even in scenarios where the prompt only comprises nouns without attributes, our method adeptly discerns the implicit taste of users toward these objects, resulting in image generation much closer to the users' actual preferences.

\section{Failure Cases}
Our Personalized PR is a first trail of personalized text-to-image prompt rewriting, where we evaluate our methods using naive yet effective methods and models.
Nevertheless, we would like to present failure cases of our Personalized PR method. While such failures do occur for some reasons, they are exceptionally rare. From which, we expect that our method's failure could help promote further studies on personalized prompt rewriting.

Figure ~\ref{fig: appen fail1} illustrate a failure case where irrelevant objects are involved into the image. We attribute the results into two reasons: the irrelevant retrieval result $\mathcal{R}_t$, and the misconception of Rewriter. For the irrelevant retrieval result, the retrieved historical prompts describe some totally different scenes, such as ``group of angels" and ``Sky City" in the third retrieved history shown in Figure ~\ref{fig: appen fail1}. For the misconception of Rewriter, the Rewriter ChatGPT ~\cite{ChatGPT} captures some objects irrelevant to ``fried wings" which is the user's preference, thus leading to this failure.

Furthermore, we visualize another failure case in Figure ~\ref{fig: appen fail2}. It happens when the Rewriter fails to acquire adequate personalized information from historical prompts. In this case, the user prefer to generate various objects without sequential connections in the historical prompts. Therefore, the Rewriter fails to capture useful information for prompt rewriting. As a result, the rewritten prompt can be likely the generation result sourcing from the ChatGPT's hallucination.

Even though there might be instances of failure, our Personalized PR poses a foundation method to align text-to-image prompts with user preferences. 
With our constructed PIP dataset, proposed Personalized PR method and the corresponding standard evaluation, we have paved a pathway of personalization in text-to-image generation and we expect to foster further research towards better user preference extraction.

\end{document}